\documentclass{article}

\PassOptionsToPackage{numbers, compress}{natbib}

\usepackage[preprint]{neurips_2025}

\usepackage[utf8]{inputenc} 
\usepackage[T1]{fontenc}    
\usepackage{hyperref}       
\usepackage{url}            
\usepackage{booktabs}       
\usepackage{amsfonts}       
\usepackage{nicefrac}       
\usepackage{microtype}      
\usepackage{xcolor}         

\usepackage{multirow} 
\usepackage{colortbl} 
\usepackage{tabularx} 
\usepackage{caption} 
\usepackage{graphicx} 
\usepackage{xcolor} 
\usepackage[utf8]{inputenc} 
\usepackage[T1]{fontenc}    
\usepackage{url}            
\usepackage{booktabs}       
\usepackage{amsfonts}       
\usepackage{nicefrac}       
\usepackage{microtype}      
\usepackage{xcolor}         
\usepackage{tabularx}
\usepackage{subcaption}
\usepackage{multicol}
\usepackage{multirow}
\usepackage{natbib}
\usepackage{tabularray}
\usepackage{amsmath}
\usepackage{amssymb}
\usepackage{siunitx}
\usepackage{colortbl}
\usepackage{graphicx}
\usepackage{pifont}
\usepackage{lipsum}
\newcommand{\Xhline}[1]{\noalign{\hrule height #1}}
\usepackage[export]{adjustbox}
\usepackage{wrapfig}
\usepackage{svg}
\usepackage{amssymb}
\usepackage{sidecap}
\usepackage{caption}
\usepackage{algorithm}
\usepackage{algorithmic}
\usepackage{array}
\usepackage{tabularx}
\usepackage[capitalize,noabbrev]{cleveref}
\usepackage{subcaption}

\newcolumntype{C}{>{\centering\arraybackslash}X}
\definecolor{cvprblue}{rgb}{0.21,0.49,0.74}
\definecolor{pltblue}{RGB}{174, 199, 232}
\definecolor{pltyellow}{HTML}{FFFACD} 
\definecolor{pltorange}{RGB}{255, 229, 204}
\definecolor{pltgreen}{RGB}{204, 229, 204}
\definecolor{pltred}{RGB}{229, 204, 204}
\definecolor{pltpurple}{RGB}{239, 218, 230}

\definecolor{tabblue}{HTML}{1f77b4}
\definecolor{taborange}{HTML}{ff7f0e}
\definecolor{tabgreen}{HTML}{2ca02c}
\definecolor{tabred}{HTML}{d62728}
\definecolor{tabpurple}{HTML}{9467bd}

\definecolor{cblue}{RGB}{173, 201, 233}
\definecolor{clblue}{RGB}{222, 234, 246}
\definecolor{corange}{RGB}{255, 152, 67}
\definecolor{lorgange}{RGB}{255, 221, 149}

\usepackage{pgfplots,gincltex}
\usepgfplotslibrary{groupplots}
\pgfplotsset{compat=1.6}
\usepackage{tikz}
\usetikzlibrary{arrows.meta}

\title{Latent Harmony: Synergistic Unified UHD Image Restoration via Latent Space Regularization and Controllable Refinement}

\author{
  \vspace{-25pt} \\
  \textbf{Yidi Liu}$^{1}$,\quad
  \textbf{Xueyang Fu}$^{1,*}$,\quad
  \textbf{Jie Huang}$^{1,\dag}$,\quad
  \textbf{Jie Xiao}$^1$,\quad
  \textbf{Dong Li}$^{1}$,\quad \\
  \textbf{Wenlong Zhang}$^{2,*}$,\quad
  \textbf{LEI BAI}$^{2}$,\quad
  \textbf{Zheng-jun Zha}$^{1}$ \\
  $^1$University of Science and Technology of China \quad\quad $^2$Shanghai AI Laboratory \\
  \texttt{\small liuyidi2023@mail.ustc.edu.cn,\quad xyfu@ustc.edu.cn} \\
  \texttt{\small $^*$ Corresponding Author},\quad $\dag$: project lead. \\
  \vspace{8pt}
}

\begin{document}

\maketitle

\begin{abstract}

Ultra-High Definition (UHD) image restoration struggles to balance computational efficiency and detail retention.
While Variational Autoencoders (VAEs) offer improved efficiency by operating in the latent space, with the Gaussian variational constraint, this compression preserves semantics but sacrifices critical high-frequency attributes specific to degradation and thus compromises reconstruction fidelity. Consequently, a VAE redesign is imperative to foster a robust semantic representation conducive to generalization and perceptual quality, while simultaneously enabling effective high-frequency information processing crucial for reconstruction fidelity.
To address this, we propose  \textit{Latent Harmony}, a two-stage framework that reinvigorates VAEs for UHD restoration by concurrently regularizing the latent space and enforcing high-frequency-aware reconstruction constraints. 
Specifically, Stage One introduces the LH-VAE, which fortifies its latent representation through visual semantic constraints and progressive degradation perturbation for enhanced semantics robustness; meanwhile, it incorporates latent equivariance to bolster its high-frequency reconstruction capabilities. 
Then, Stage Two facilitates joint training of this refined VAE with a dedicated restoration model.
This stage integrates High-Frequency Low-Rank Adaptation (HF-LoRA), featuring two distinct modules: an encoder LoRA, guided by a fidelity-oriented high-frequency alignment loss, tailored for the precise extraction of authentic details from degradation-sensitive high-frequency components; and a decoder LoRA, driven by a perception-oriented loss, designed to synthesize perceptually superior textures. These LoRA modules are meticulously trained via alternating optimization with selective gradient propagation to preserve the integrity of the pre-trained latent structure. This methodology culminates in a flexible fidelity-perception trade-off at inference, managed by an adjustable parameter 
$\alpha$.
Extensive experiments demonstrate that \textit{Latent Harmony} effectively balances perceptual and reconstructive objectives with efficiency, achieving superior restoration performance across diverse UHD and standard-resolution  scenarios. The code will be available at \url{https://github.com/lyd-2022/Latent-Harmony}.

\end{abstract}

\begin{figure*}[t!]
	\centering
	\includegraphics[width=1\textwidth]{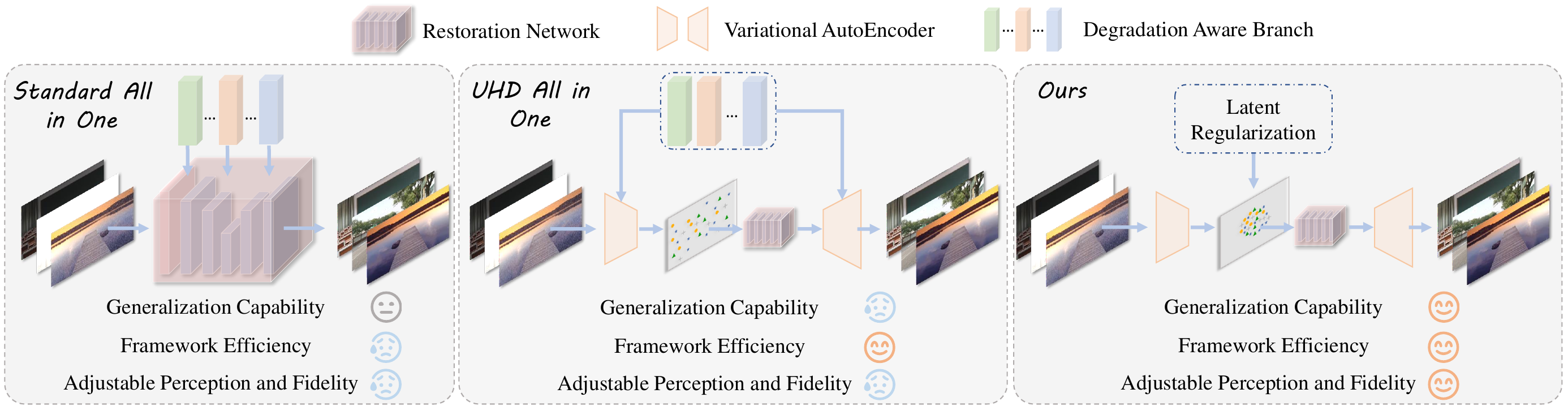} %
	\caption{Comparison with existing mainstream methods.Our method outperforms existing standard and UHD all-in-one approaches by leveraging latent regularization, achieving superior efficiency and generalization without requiring degradation-aware branches, while enabling adjustable fidelity and perceptual quality during inference.}
	\label{fig:Introduction}
    \vspace{-2em}
\end{figure*}

\section{Introduction}

Image restoration~\cite{Zamir2021Restormer,chen2022simple,huang2022deep,yu2022frequency,WhenFast,yang2024unleashing,xu2025adaptive} aims to recover high-quality images from their low-quality degraded versions with semantic recovery and detail reconstruction, which often struggles to handle unknown corruptions in real-world scenarios.
To address this, all-in-one image restoration methods~\cite{cao2024hairhypernetworksbasedallinoneimage,potlapalli2023promptir,cui2024adair} develop a single model for multiple degradation types, striving for broad generalization. However, they often face computational efficiency challenges, particularly at high resolutions.
In contrast, \textbf{Ultra-High Definition (UHD) image restoration}~\cite{liu2025uhd,uhdhaze,zheng2022uhd,liu2025dreamuhd,xiao2022singleuhdimagedehazing,wang2024uhdformer} specifically targets the immense data scale and intricate detail preservation required for 4K images. 
To meet complex scenes with various degradations and high resolution,  \textbf{UHD all-in-one image restoration}~\cite{liu2025uhd} is developed based on the above works, amplifying the challenges of robust generalization, extreme computational efficiency, and meticulous detail fidelity.

To enhance efficiency, existing methods employ Variational Autoencoders (VAEs) to migrate the core restoration process to a lower-dimensional, compact latent space~\cite{liu2025dreamuhd,liu2024cosae,liu2025uhd,liu2025decouple}.
This approach allows downstream de-degradation networks to operate on significantly smaller latent space, and finally decoding step back to the original space to reconstruct fidelity details while preserving semantics.

However,  directly applying VAEs to complex all-in-one UHD image restoration tasks reveals inherent limitations. 
The VAE's typically achieve compression through Gaussian variational inference, excels at preserving global robust semantic representations~\cite{vae}. 
Yet, this mechanism often leads to the loss of critical high-frequency attributes intertwined with degradation characteristics, restricting VAE's ability for high-fidelity reconstruction with the loss of fine details and textures affected by various corruptions.
Therefore, a fundamental challenge arises: how to redesign VAE mechanisms to effectively trade-off two crucial properties: (1) extracting degradation-robust semantic representations that ensure good generalization and perceptual quality, and (2) ensuring that these latent representations, upon reconstruction to original pixel space, can adequately process and represent degradation-related high-frequency information for high reconstruction fidelity.

To address this challenge, this paper proposes \textit{Latent Harmony} (LH), a novel two-stage synergistic framework. The core idea of the LH framework is to enable the latent representation to possess both strong semantic robustness and high reconstruction capability by simultaneously regularizing the latent space and imposing high-frequency-related reconstruction constraints.

The first stage introduces the LH-VAE as the foundation for all-in-one UHD image restoration. 
In its encoding process, building upon the VAE's original Gaussian distribution constraint for latent, the LH-VAE further incorporates progressive degradation perturbation and visual semantic constraints to enhance latent semantic robustness. 
Concurrently, during its decoding process, latent space equivariance constraints are introduced to improve the latent representation's intrinsic ability to reconstruct high-frequency components. 
This stage aims to construct a generalized latent space resilient to various degradations and possessing a more balanced frequency characteristic for reconstruction. 

The second stage is built on the above VAE. This stage involves joint training with a restoration model, addressing the VAE co-optimization and the perception-fidelity balance. 
After initially training a latent space restoration network with a fixed LH-VAE, an innovative high-frequency-guided Low-Rank Adaptation (HF-LoRA) fine-tuning mechanism is introduced. 
To manage degradation-sensitive high-frequency information and enhance fidelity, Fidelity-oriented HF-LoRA (FHF-LoRA) is introduced into the encoder, guided by a high-frequency alignment loss that aligns with the restoration model. 
Meanwhile, to enhance the perceptual quality of reconstructed output, perception-oriented HF-LoRA (PHF-LoRA) is incorporated into the decoder guided by a high-frequency perception loss. 
These LoRA modules are trained via alternating optimization with the corresponding losses, thereby protecting the pre-trained VAE's structure from potentially disruptive gradients.
Furthermore, the framework allows users to flexibly balance the fidelity-oriented and perception-oriented high-frequency contributions in the final output via an adjustable parameter $\alpha$ during inference. 

The main contributions of this paper include: 
 \begin{itemize}
 \item We construct a new Latent Harmony two-stage framework, which systematically addresses the multiple trade-off challenges in UHD all-in-one image restoration. 
 \item We design a new latent space regularization strategy that combines progressive degradation, semantic alignment, and equivariance constraints to construct a high-quality generalized VAE latent space.
 \item  We propose a pioneering high-frequency-guided LoRA fine-tuning paradigm that optimizes encoder LoRA for fidelity and decoder LoRA for perception, achieving a synergistic solution for enhanced performance, VAE structural integrity, and controllable output characteristics.
 \item Extensive experiments demonstrate the superiority of the proposed framework across various UHD and standard-resolution degradation scenarios.
 \end{itemize}

\section{Related Work}

\subsection{UHD and All-in-One Image Restoration}

\textbf{Ultra-High Definition (UHD) image restoration} focuses on recovering high-fidelity images from low-quality UHD observations~\citep{liu2025dreamuhd,xiao2022singleuhdimagedehazing,wang2024uhdformer,yu2023learning,yu2024empowering,chen2025ultra}. This task poses significant challenges due to the substantial computational overhead of processing vast data volumes and the stringent requirement to preserve fine high-frequency details. Direct application of deep learning models in pixel space is computationally prohibitive for UHD images. To address this, a prevalent approach is the downsample-enhance-upsample paradigm, where UHD images are downsampled, processed, and subsequently upsampled. For example, UHDFour~\citep{Li2023ICLR} enables full-resolution inference on edge devices through 8x downsampling, UHDformer~\citep{wang2024uhdformer} leverages high-resolution features to guide low-resolution restoration, and UDR-Mixer~\citep{chen2024towards} employs frequency feature modulation to enhance spatial feature recovery at lower resolutions. However, this downsampling strategy inevitably incurs information loss, particularly detrimental to UHD images with intricate textures, which subsequent processes struggle to fully recover, thus capping restoration quality.In parallel, an alternative efficiency-driven approach utilizes latent space models, notably Variational Autoencoders (VAEs), to shift the restoration process into a lower-dimensional latent space. For instance, DreamUHD~\citep{liu2025dreamuhd} integrates a VAE framework with frequency augmentation and high-frequency injection to manage UHD details, while CD²-VAE~\citep{liu2025decouple} employs active feature decoupling and a reversible fusion network within a VAE structure to balance background consistency and degradation removal. These efforts highlight the pivotal role of meticulously designed latent space models in UHD restoration. Nevertheless, enhancing latent representation capacity to accommodate more complex degradations while mitigating inherent VAE limitations—such as the trade-off between generalization and reconstruction fidelity—remains a critical research frontier.

\textbf{All-in-One image restoration} aims to devise a unified model capable of addressing diverse, mixed, or unknown degradation types, necessitating exceptional generalization and adaptability~\cite{AirNet,cui2024adair,potlapalli2023promptir,MIOIR,yao2024neural,yu2024multiexpertadaptiveselectiontaskbalancing}. Conventional methods typically design models tailored to specific degradations, a strategy impractical for real-world scenarios characterized by complex and variable degradation patterns. All-in-One models must contend with challenges such as managing degradation heterogeneity, mitigating conflicts among restoration sub-tasks, and achieving awareness of unknown degradations. Prevailing approaches predominantly adopt a strategy that combines a degradation-aware branch with an image restoration backbone. The degradation-aware branch is typically designed based on Mixture-of-Experts (MoE)~\cite{lin2024moe,yu2024multiexpertadaptiveselectiontaskbalancing} or Prompting~\cite{gao2023prompt,ma2023prores,li2023promptinpromptlearninguniversalimage}, while the image restoration backbone often employs established architectures such as Restormer or NAFNet. For instance, PromptIR~\cite{potlapalli2023promptir} pioneered the introduction of a prompt-based degradation-aware branch into the all-in-one image restoration task, enhancing the model's adaptability to diverse degradations. Similarly, MoCE-IR~\cite{zamfir2024complexityexperts}, through its MoE design, enables specialized handling of different degradation inputs, further bolstering performance. Although these methods have demonstrated superior performance, their overall network efficiency remains relatively low (a challenge pertinent to the general 'Restoration Network' paradigms, as conceptually outlined in the leftmost panel of main text Figure 1 ), making full-resolution inference on high-resolution images challenging on consumer-grade GPUs, which consequently limits their practical applicability.

\subsection{Variational Autoencoders and Latent Space Optimization}

\textbf{Variational Autoencoders} (VAEs) are extensively utilized in image restoration owing to their encoder-decoder architecture, which maps images into a low-dimensional latent space for reconstruction. The VAE's learning objective, the Evidence Lower Bound (ELBO), balances reconstruction fidelity against latent space regularity through the KL divergence term, constraining the latent distribution to a prior. This inherent trade-off significantly influences restoration quality: excessive regularization may yield insufficient latent information, resulting in blurry reconstructions, whereas prioritizing reconstruction can compromise latent space structure and generalization. VAEs have proven effective in tasks such as denoising, deblurring, and super-resolution, and they underpin Latent Diffusion Models (LDMs), where latent space quality dictates performance ceilings. However, the information compression inherent to VAEs often leads to the loss of high-frequency details, a pressing concern for UHD restoration. Efforts like FA-VAE~\citep{favae2023cvpr} incorporate frequency-complementary modules and dynamic spectral losses to bolster high-frequency reconstruction, while Wavelet-VAE~\citep{kiruluta2025wavelet} and LiteVAE~\citep{sadat2024litevae} leverage wavelet transforms to improve capture and recovery of high-frequency components. These developments underscore the necessity of tailoring VAEs to preserve high-frequency information for superior restoration outcomes.

\textbf{latent space regularization} To address the shortcomings of standard VAEs and enhance the quality and generalization of latent representations, researchers have explored diverse latent space regularization strategies. Beyond tuning the KL divergence weight as in $\beta$-VAE~\citep{higgins2017beta}, techniques include contrastive learning to boost discriminability (e.g., Hi-CDL in CD²-VAE~\citep{liu2025decouple}), geometric regularization~\cite{eqvae,zhou2025aliasfreelatentdiffusionmodelsimproving} to shape the latent manifold, and diffusion-based decoders to elevate generation quality (e.g., $\epsilon$-VAE~\citep{zhao2025epsilonvaedenoisingvisualdecoding}). These approaches aim to render latent representations more resilient to transformations such as degradations or geometric shifts. For instance, aligning VAE latent variables with features from robust pre-trained vision models like DINOv2~\citep{oquab2024dinov}—as seen in VAVAE~\citep{yao2025vavae}—injects valuable semantic priors, enhancing robustness and generalization. Furthermore, works such as REPA~\citep{yu2025repa} and REPA-E~\citep{leng2025repaeunlockingvaeendtoend} investigate feature alignment or alignment losses to optimize training, including end-to-end joint training of VAEs and LDMs, which also refines latent space structure. These insights suggest that integrating multiple regularization strategies, particularly by leveraging external priors and internal structural constraints, offers a promising avenue for crafting a latent space optimized for All-in-One UHD image restoration—a foundational principle of our Latent Harmony framework’s initial stage.

\section{Motivation}
\label{Motivation}

\begin{figure*}[t!]
	\centering
	\includegraphics[width=1\textwidth]{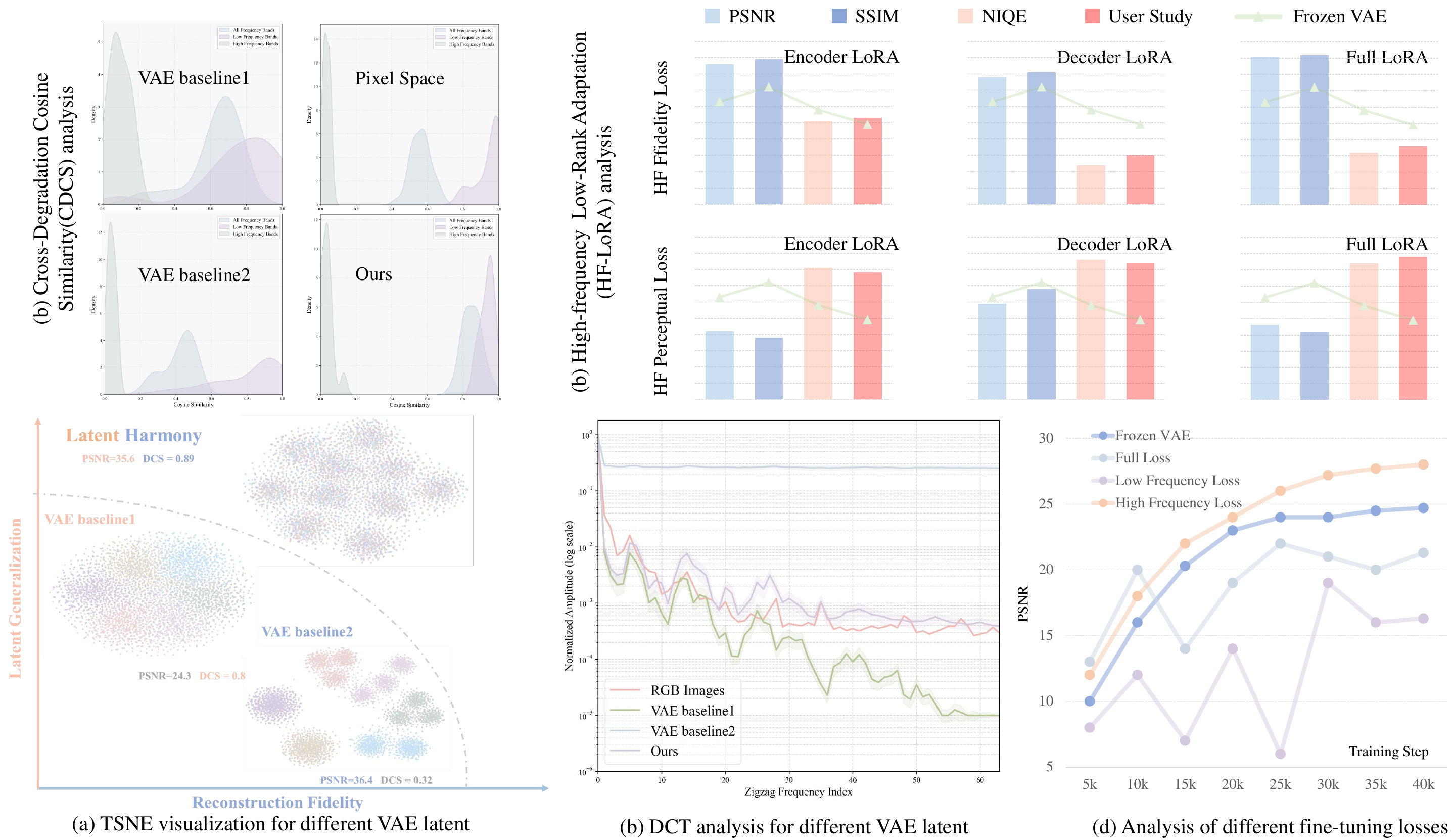} %
	\caption{Motivation Analysis. (a) t-SNE visualization of VAE latents under diverse degradations, showing Baseline2's degradation-sensitive clustering versus our method's semantic clustering. (b) Cross-degradation cosine similarity (CDCS) analysis, with higher CDCS in high-frequency bands. (c) DCT spectral analysis, revealing Baseline1's low high-frequency components and Baseline2's elevated components, indicating a reconstruction-generalization trade-off via latent high-frequency proportions. (d) Fine-tuning loss comparison, highlighting stable downstream gains with high-frequency loss. (e) HF-LoRA experiments, demonstrating optimal fidelity and perceptual gains from encoder (fidelity loss) and decoder (perceptual loss) fine-tuning.(Note: All metrics in (e) are normalized to a positive scale, where higher values indicate better performance)}
\label{fig:motivation}
    \vspace{-1em}
\end{figure*}

\subsection{Latent Space Representation: Generalization vs. Reconstruction}
\label{sec:M1}
To inform our VAE's design for UHD all-in-one image restoration, we conducted experiments comparing two VAE baselines: standard VAE (baseline1) and VAE with enhanced reconstruction (baseline2, Appendix). 
By analyzing the Cross-Degradation Cosine Similarity (CDCS) of their latent representations for diverse degraded inputs in Fig.~\ref{fig:motivation}(b), we find that stronger reconstruction (baseline2) yields lower latent CDCS (i.e., more diverse among degradations), even below the input's pixel-space CDCS. 
The t-SNE in Fig.~\ref{fig:motivation}(a) depicts a similar phenomenon that stronger reconstruction promotes degradation-driven clustering in latent space, undermining content-based organization.
\textbf{These results suggest that enhancing VAE's reconstruction capability makes the latent space more sensitive to input degradations}, posing challenges for downstream restoration networks.

We then further perform frequency-domain analysis for latent representations in Fig.~\ref{fig:motivation}(b), depicting
that high-frequency exhibits low CDCS and low-frequency depicts higher CDCS.
Subsequently, a more detailed frequency analysis is performed for both pixel and latent spaces, referring to ~\cite{skorokhodov2025improvingdiffusabilityautoencoders}.
Results in Fig.~\cref{fig:motivation}(c) indicate that VAEs with stronger reconstruction capabilities (baseline2) tend to encode a significantly higher proportion of high-frequency components in their latent space compared to the pixel space. 
\textbf{This implies that a VAE with strong reconstruction ability encodes ample high-frequency information to manage the challenging task of decoding consistent fine details.}

Moreover, we find that the high-frequency components inherently exhibit low CDCS (Fig.~\ref{fig:motivation}(b)) among degradations in latent.
Based on the above observations, \textbf{an excessive proportion of high frequencies is beneficial for detailed reconstruction but compromises the latent space's robustness to varied degradations.}  
Therefore, our \textbf{first core impetus} is to constrain the latent space to a more moderate high-frequency proportion, balancing latent representation robust to degradations (high CDCS in latent) for generalization and has good reconstruction capabilities (low CDCS in latent).

\vspace{-0.5em}
\subsection{VAE Co-optimization: Downstream Adaptation vs. Structural Preservation}
\label{sec:M2}
\vspace{-0.5em}
Following the pre-trained VAE with initial generalization capabilities, a critical question arises: should this VAE remain fixed for downstream restoration tasks, or can downstream restoration supervision signals further adapt the pre-trained VAE to improve the final restoration performance?

While co-optimizing the VAE with the downstream restoration network theoretically promises improved overall performance, direct joint optimization is fraught with risks. 
Drawing parallels from Latent Diffusion Models (LDMs), where directly applying the main diffusion loss to the VAE can be detrimental~\cite{leng2025repaeunlockingvaeendtoend}.
We conceptualize an observational experiment, the pretrained VAE is fine-tuned by backpropagating the restoration loss $L_{Res}$ from the downstream network $R_\theta$. 
As illustrated in Fig.~\ref{fig:motivation}(d), unfreezing the VAE yields faster initial PSNR gains compared to its frozen version; however, continued training often results in performance oscillations. 
This phenomenon can be attributed to the direct optimization pressure of $L_{Res}$, which compels the encoder to prematurely and aggressively remove degradations from the input. Consequently, \textbf{this direct "encoder-latent-restoration-decoder" optimized paradigm tends to devolve into a simplified bottleneck structure geared toward direct pixel-level restoration, thereby disrupting the learned latent space structure.}
In contrast, \textbf{while a frozen VAE ensures training stability, its lack of adaptation to the restoration task leads to a performance bottleneck that cannot be overcome in later training stages.}

The success of REPA-E~\cite{leng2025repaeunlockingvaeendtoend} achieves effective co-optimization using representation alignment, suggesting a path. 
In our restoration task, while Stage 1 for VAE's generalization might reduce some high-frequency components, the final reconstruction quality depends on recovering these details. 
This motivates using high-frequency information alignment as a "bridge" loss for co-optimization.

In our restoration task, as discussed in Section~\ref{sec:M1}, enhanced generalization often sacrifices some high-frequency information, while low-frequency components are effectively encoded within a structured latent space. 
\textbf{To balance the preservation of the pre-trained latent structure in VAE with performance gains in downstream restoration tasks during joint optimization, we focus the optimization objective on high-frequency components, employing high-frequency information alignment as a "bridge" loss for co-optimization. }
As shown in Fig.~\ref{fig:motivation}(d), backpropagating the high-frequency loss to update the VAE maintains training stability while overcoming performance bottlenecks of restoration. 
Conversely, using a low-frequency alignment loss leads to training instability. 
Thus, our \textbf{second core impetus} is to introduce high-frequency alignment loss as a bridge for joint optimizing VAE and restoration network, preserving the VAE’s pre-trained, highly generalizable representations while achieving further performance improvements driven by downstream tasks.

\subsection{High-Frequency Restoration From Latent: Perception vs. Fidelity}

To address the need for task-specific VAE adaptation through high-frequency guidance (see Section~\ref{sec:M2}), we focus on designing the high-frequency alignment loss and analyzing its impact on output quality. 
High-frequency detail recovery entails a trade-off between perceptual quality and fidelity. 
Pixel-level losses ($L_{pix}$), typically formulated as Maximum Likelihood Estimation, minimize both Systematic Effect (SE) that affects fidelity via regressable components like edges, and Variance Effect (VE) that influences perception through non-regressable textures~\cite{lee2025auto}. Minimizing $L_{pix}$, including its high-frequency component, suppresses SE and VE, yielding high PSNR but perceptually flat outputs due to reduced VE. This is expressed as:
\begin{equation}
\min_{\hat{y}} \left\{ \underbrace{\mathbb{E}_{y} \left[ \mathcal{L}(y, \mu_{\hat{y}}) - \mathcal{L}(y, \mu_y) \right]}_{\text{SE: LF + regressable HF}} + \underbrace{\mathbb{E}_{y, \hat{y}} \left[ \mathcal{L}(y, \hat{y}) - \mathcal{L}(y, \mu_{\hat{y}}) \right]}_{\text{VE: non-regressable HF}} \right\},
\end{equation}
where $\mathcal{L}$ is a symmetric loss, $y \sim p(y|x)$, $\hat{y}$ estimates $y$, and $\mu_y$, $\mu_{\hat{y}}$ are their respective means.

\textbf{(a) Fidelity-Oriented High-Frequency Restoration:} This approach prioritizes the faithful extraction or disentanglement of authentic high-frequency components from the input signal, aligning with the ground truth $I_{\text{clean}}$. It emphasizes the "traceability" of high-frequency details, aiming to closely match the ground truth and achieve high fidelity metrics. 
However, its efficacy is constrained by the availability of residual high-frequency information in the input. 
Moreover, the suppression of variance effects (VE) can result in monotonous textures, thereby limiting overall perceptual quality.

\textbf{(b) Perception-Oriented High-Frequency Generation:} This strategy focuses on generating visually natural high-frequency details, which may not precisely map to the input signal but rely heavily on learned priors about natural images' appearance high frequencies. 
It prioritizes the visual "plausibility" of high-frequency information, aiming to preserve or shape VE for enhanced visual quality. However, it may introduce structural inaccuracies or hallucinations and compromise fidelity.

The analysis reveals that mechanisms targeting structural fidelity (SE reduction) and texture perception (VE preservation/shaping) inherently pursue distinct optimization objectives. 
To address these coupled objectives, we introduce two independent Low-Rank Adaptation (LoRA) modules.
Specifically, we fine-tune the VAE's encoder, decoder, or both using fidelity-oriented and perception-oriented high-frequency losses, respectively, and evaluate their impact on fidelity and perceptual metrics. 
As shown in Fig.~\ref{fig:motivation}(e), \textbf{fine-tuning the VAE's encoder with the fidelity loss enhances fidelity metrics with minimal perceptual quality degradation, while fine-tuning the VAE's decoder with the perceptual loss improves perceptual metrics at a modest cost to fidelity.}

This insight forms \textbf{our third core impetus}: within a high-frequency-alignment-based VAE fine-tuning framework, we propose differentiated mechanisms. 
One mechanism focuses on faithfully extracting and aligning regressable high-frequency components to enhance fidelity in VAE's encoder, while the other concentrates on generating perceptually superior non-regressable high-frequency details to improve perceptual quality in VAE's decoder. 
We posit that balancing these independently guided mechanisms can effectively synergize fidelity and perception.

\vspace{-0.5em}
\section{ Methodology}
\label{Methodology}
\vspace{-0.5em}

\begin{figure*}[t!]
	\centering
	\includegraphics[width=1\textwidth]{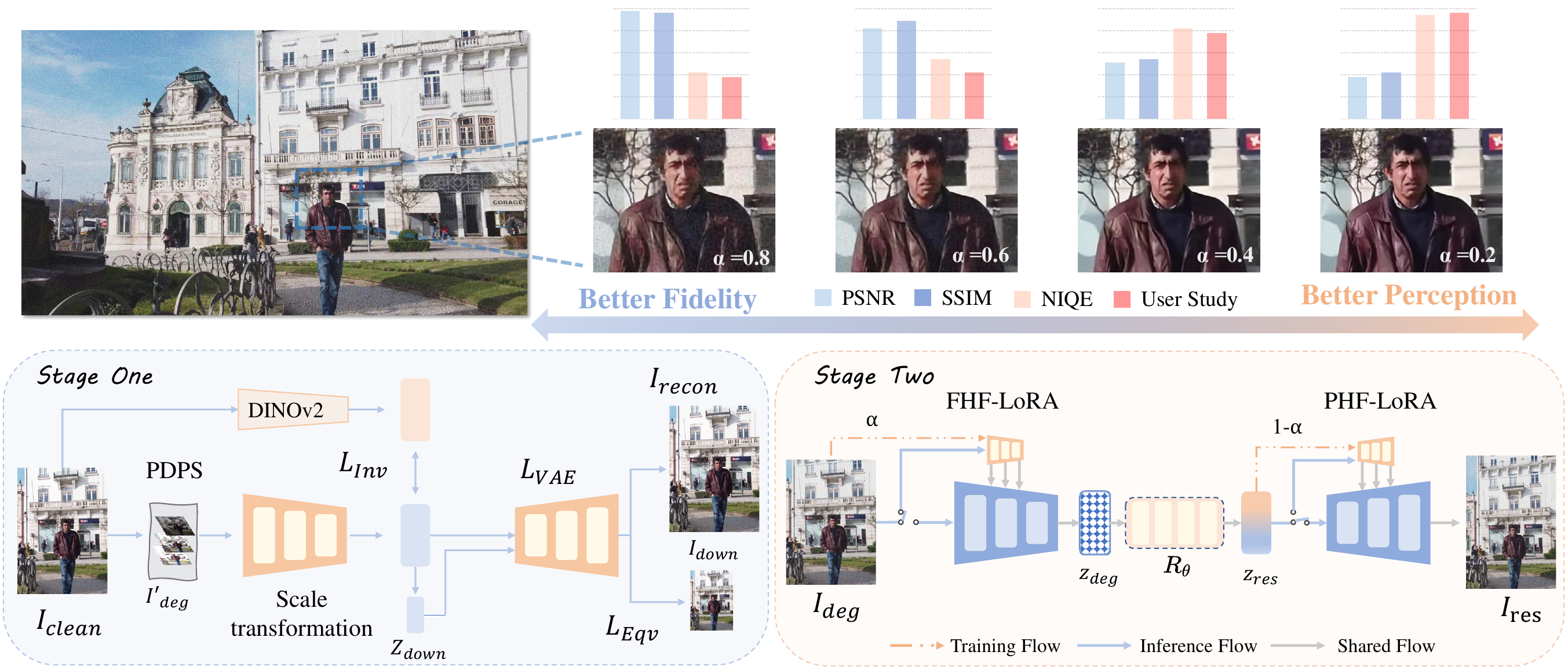} %
	\caption{Framework Overview. Stage 1: LH-VAE training employs progressive degradation perturbation, degradation-invariant visual semantic loss $L_{INV}$, and latent space equivariance loss $L_{Eqv}$ to construct a robust, generalizable latent space. Stage 2: Latent space restoration leverages $R_\theta$ and high-frequency-guided LoRA fine-tuning, with Fidelity-oriented HF-LoRA (FHF-LoRA) for the encoder and Perception-oriented HF-LoRA (PHF-LoRA) for the decoder, enabling adjustable fidelity and perceptual quality via parameter $\alpha$ during inference. Results of $\alpha$ tuning are shown in the upper panel, with metrics normalized positively, where higher values indicate better performance.}
        \vspace{-1em}
	\label{fig:framework}
\end{figure*}

Building on~\cref{Motivation}, this chapter details our proposed Latent Harmony two-stage synergistic framework. 
This framework, through latent space regularization in the first stage and high-frequency guided controllable refinement in the second stage, addresses the trade-offs between: (1) latent space generalization and reconstruction fidelity, (2) VAE co-optimization with downstream tasks versus structural preservation, and (3) the final output's perceptual quality versus fidelity.

\subsection{Stage One: Constructing a Generalizable Latent Space Representation}
\vspace{-0.5em}

The objective of this stage is to train a Variational Autoencoder (VAE) comprising an encoder $E_\phi$ and a decoder $D_\psi$, such that it learns a latent space $Z$ robust to various degradations. The base training follows the standard VAE objective, optimizing a loss $L_{VAE}$ that includes an L1 reconstruction loss on clean images $I_{clean}$ and a KL divergence regularizer:

\begin{equation}
    L_{VAE} = \left\| D_\psi(E_\phi(I_{clean})) - I_{clean} \right\|_1 + \lambda_{KL} \cdot \text{KL}\left[q_\phi(z \mid I_{clean}) \ \| \ p(z)\right]
\end{equation}

To counteract the standard VAE latent space's sensitivity to degradations, particularly in high-frequency components , we introduce a progressive degradation perturbation strategy(PDPS). During training, increasingly severe degradations are applied to $I_{\text{clean}}$ over time $t$. This perturbation is probabilistic and can take one of three forms: no perturbation, synthetic degradation, or interpolation with a paired real degraded image $I_{\text{deg}}$. The severity of synthetic degradations is controlled by an increasing function $\text{sev}(t)$, and the interpolation with $I_{\text{deg}}$ is controlled by an increasing coefficient $\beta(t)$. Formally, the perturbed image $I'_{\text{deg}}$ is generated as:

\begin{equation} 
    I'_{\text{deg}} =
    \begin{cases}
    I_{\text{clean}} & \text{with probability } p_0 \\
    \text{SynthDeg}(I_{\text{clean}}, \text{sev}(t)) & \text{with probability } p_1 \\
    (1 - \beta(t)) I_{\text{clean}} + \beta(t) I_{\text{deg}} & \text{with probability } p_2
    \end{cases}
\end{equation}
where $p_0 + p_1 + p_2 = 1$ . $\text{SynthDeg}(I, \text{sev}(t))$ applies a randomly selected set of synthetic degradations (e.g., Gaussian noise, blur, JPEG compression) to image $I$, with their severity controlled by $\text{sev}(t)$, a monotonically increasing function of $t$. The interpolation coefficient $\beta(t) \in [0,1]$ is also a monotonically increasing function of $t$, signifying a progressively stronger influence of the paired degraded image. This progressive approach ensures learning stability.

On this basis, two key regularization losses are incorporated. The degradation invariance visual semantic loss $L_{INV}$ leverages semantic features $f_{VFM} = \text{VFM}(I_{clean})$ extracted from a pre-trained DINOv2~\cite{oquab2024dinov} model as a reference, enforcing the encoder $E_\phi$ to align the encoding $z'_{deg} = E_\phi(I'_{deg})$ of the perturbed image with this reference, learning a degradation-invariant content representation:
\begin{equation}
    L_{Inv} = d(z'_{deg}, f_{VFM})
\end{equation}

where $d(\cdot, \cdot)$ denotes a distance metric in the feature space. Additionally, the latent space equivariance loss $L_{Eqv}$ constrains the consistency between the decoded result of a randomly downsampled latent encoding $z_{down} = \text{Down}_s(z_{clean})$ and the corresponding downsampled image $I_{down} = \text{Down}_s(I_{clean})$, enhancing scale robustness and reducing reliance on high-frequency components:
\begin{equation}
    L_{Eqv}  = \left\| D_\psi(z_{down}) - I_{down} \right\|_1
\end{equation}

The joint optimization objective for this stage combines these terms as:
\begin{equation}
    L_{Stage1} = L_{VAE} + \lambda_{Inv} L_{Inv} + \lambda_{Eqv} L_{Eqv}
\end{equation}

Optimizing this objective yields VAE parameters $(\phi^*, \psi^*)$ that define a latent space exhibiting stronger cross-degradation consistency and more balanced frequency characteristics, establishing a generalizable foundation for subsequent restoration, albeit potentially at the cost of reducing high-frequency information useful for reconstruction.

\subsection{Stage Two: High-Frequency Guided Controllable Low-Rank Adaptation}

This stage aims to achieve high-quality image restoration using the generalizable latent space $(\phi^*, \psi^*)$ from Stage One, while compensating for lost high-frequency details and providing controllability over the final output. Initially, a latent space restoration network $R_\theta$ is introduced, which processes the encoded degraded latent $z_{deg} = E_{\phi^*}(I_{deg})$ to predict a restored latent $z_{res} = R_\theta(z_{deg})$. This network is trained solely with a standard restoration loss, keeping VAE's parameters $(\phi^*, \psi^*)$ frozen:
\begin{equation}
    L_{Res} = \left\| D_{\psi^*}(z_{res}) - I_{clean} \right\|_1
\end{equation}

Gradients update only the parameters $\theta$ via $\theta \leftarrow \theta - \eta \nabla_\theta L_{Res}$. 

Subsequently, to finely restore high-frequency information without compromising the acquired generalization,  high-frequency-guided Low-Rank Adaptation (HF-LoRA) fine-tuning is applied to the pre-trained VAE. Low-rank updates $\Delta\phi_{LoRA}$ and $\Delta\psi_{LoRA}$ are introduced to the base parameters $\phi^*$ and $\psi^*$, such that $\phi = \phi^* + \Delta\phi_{LoRA}$ and $\psi = \psi^* + \Delta\psi_{LoRA}$. The LoRA parameters $\theta_{LoRA} = \{\Delta\phi_{LoRA}, \Delta\psi_{LoRA}\}$ are optimized solely by a specific high-frequency alignment loss $L_{HF}$, decoupled from the main restoration loss $L_{Res}$, to preserve the latent space structure learned in Stage One. We design Fidelity-oriented HF-LoRA (FHF-LoRA) for the encoder, guided by a high-frequency alignment loss to enhance fidelity, and perception-oriented HF-LoRA (PHF-LoRA) for the decoder, guided by a high-frequency perception loss to improve perceptual quality, with both modules trained using an alternating optimization strategy.
\begin{figure*}[t!]
	\centering
	\includegraphics[width=1\textwidth]{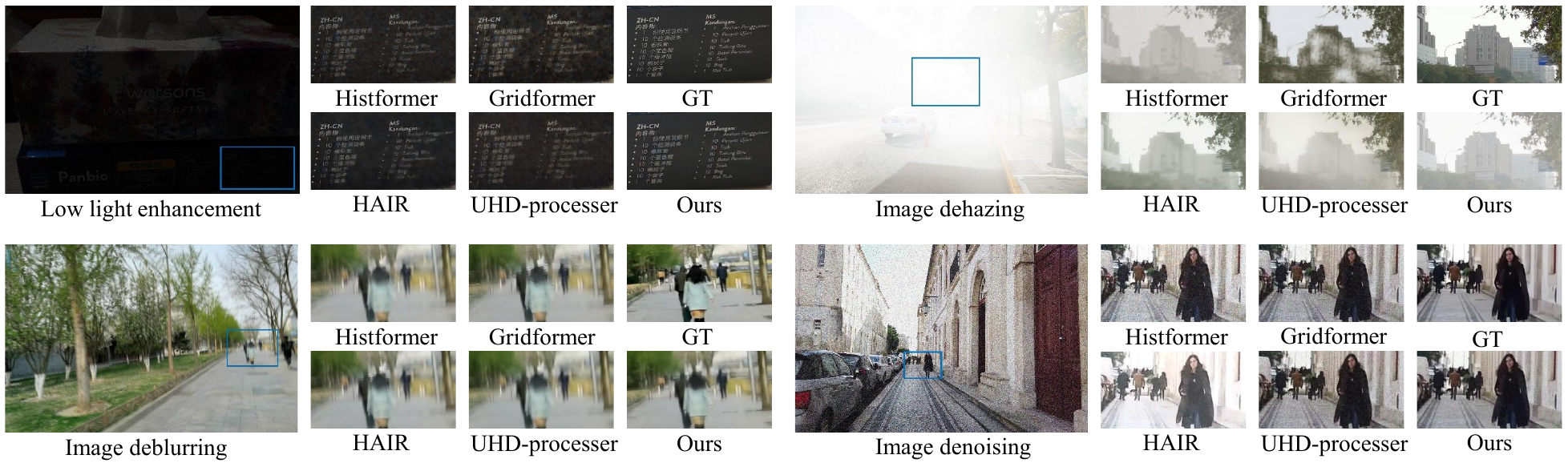} %
	\caption{Visual results for four types of degradation removal with other all-in-one methods. }
	\label{fig:visual}
\end{figure*}

When optimizing the FHF-LoRA ($\Delta\phi_{LoRA}$), the decoder uses its frozen base parameters $\psi^*$, with the objective being a high-frequency fidelity loss $L_{HF_{Fid}}$ that extracts high-frequency structures from the degraded input consistent with the ground truth:
\begin{equation}
    L_{HF_{Fid}} = \left\| \text{HF}\left(D_{\psi^*}(E_{\phi^* + \Delta\phi_{LoRA}}(I_{deg}))\right) - \text{HF}(I_{clean}) \right\|_1
\end{equation}

When optimizing the PHF-LoRA ($\Delta\psi_{LoRA}$), the encoder uses its frozen base parameters $\phi^*$, with the objective being a high-frequency perceptual loss $L_{HF_{Perc}}$ to generate visually natural and sharp high-frequency textures, enhancing perceptual quality. This loss is implemented as a GAN-based loss $L_{HF_{GAN}}$, minimizing the adversarial loss for the generator (decoder LoRA) to deceive a high-frequency discriminator $D_{HF}$. The discriminator $D_{HF}$ is optimized by adversarial loss:
\begin{equation}
    L_{HF_{GAN}} = -\mathbb{E}_{I_{deg}} \left[ \log D_{HF} \left( \text{HF}\left(D_{\psi^* + \Delta\psi_{LoRA}}(R_\theta(E_{\phi^*}(I_{deg}))) \right) \right) \right]
\end{equation}


\begin{table*}[!t]
    \centering
    \tiny
    \fboxsep0.75pt
    \setlength\tabcolsep{3pt}
    \caption{\textit{Comparison to state-of-the-art on four degradations.} PSNR (dB, $\uparrow$), \colorbox{pltorange!50}{SSIM ($\uparrow$)}, and \colorbox{pltyellow!50}{LPIPS ($\downarrow$)}, and FS represents full-size 4K image inference. FLOPs are computed for an input size of 256$\times$256. \textcolor{tabred}{\textbf{Best}} and \textcolor{tabblue}{\textbf{second best}} performances are highlighted.}
    \label{tab:exp:4deg}
    \begin{tabularx}{0.97\textwidth}{Xc*{17}{c}}
    \Xhline{1pt}
     \multirow{2}{*}{Method} & \multirow{2}{*}{FS} & \multirow{2}{*}{\centering FLOPs} & \multirow{2}{*}{\centering Params.}  
     & \multicolumn{2}{c}{\textit{Low Light}} & \multicolumn{2}{c}{\textit{Deblurring}} & \multicolumn{2}{c}{\textit{Dehazing}} & \multicolumn{6}{c}{\textit{Denoising}} 
     & \multicolumn{3}{c}{\multirow{2}{*}{\centering Average}} \\
     \cmidrule(lr){5-6} \cmidrule(lr){7-8} \cmidrule(lr){9-10} \cmidrule(lr){11-16}
     &&&& \multicolumn{2}{c}{UHD-LL} & \multicolumn{2}{c}{UHD-blur} & \multicolumn{2}{c}{UHD-haze} & \multicolumn{2}{c}{UHDN\textsubscript{$\sigma$=15}} & \multicolumn{2}{c}{UHDN\textsubscript{$\sigma$=25}} & \multicolumn{2}{c}{UHDN\textsubscript{$\sigma$=50}} \\
     \midrule
        AIRNet~\cite{AirNet} & \ding{55} & 301G & {9M} & 19.24 & \cellcolor{pltorange!50}{.809} & 21.89 &  \cellcolor{pltorange!50}{.757} & 18.37 &  \cellcolor{pltorange!50}{.812} & 21.33 &  \cellcolor{pltorange!50}{.887} & 20.78 &  \cellcolor{pltorange!50}{.784} & 18.79 &  \cellcolor{pltorange!50}{.475} & 20.07 &  \cellcolor{pltorange!50}{.754} & \cellcolor{pltyellow!50}{.2843} \\
        IDR~\cite{zhang2023ingredient} & \ding{55} & 88G & 15.3M & 23.12 & \cellcolor{pltorange!50}{.910} & 24.67 &  \cellcolor{pltorange!50}{.793} & 19.12 &  \cellcolor{pltorange!50}{.768} & 27.48 &  \cellcolor{pltorange!50}{.912} & 25.86 &  \cellcolor{pltorange!50}{.872} & 24.57 &  \cellcolor{pltorange!50}{.654} & 24.14 &  \cellcolor{pltorange!50}{.822} & \cellcolor{pltyellow!50}{.2684} \\
        PromptIR~\cite{potlapalli2023promptir} & \ding{55} & 158G & 33M & 23.44 & \cellcolor{pltorange!50}{.902} & 25.77 & \cellcolor{pltorange!50}{.782} & 19.97 & \cellcolor{pltorange!50}{.727} & 28.43 & \cellcolor{pltorange!50}{.924} & 26.74 & \cellcolor{pltorange!50}{.898} & 23.72 & \cellcolor{pltorange!50}{.584} & 24.68 & \cellcolor{pltorange!50}{.803} & \cellcolor{pltyellow!50}{.2571} \\
        CAPTNet~\cite{gao2023prompt} & \ding{55} & 25G & 24.3M & 23.96 & \cellcolor{pltorange!50}{.920} & 26.11 & \cellcolor{pltorange!50}{.798} & 19.46 & \cellcolor{pltorange!50}{.868} & 25.58 & \cellcolor{pltorange!50}{.865} & 23.24 & \cellcolor{pltorange!50}{.884} & 21.98 & \cellcolor{pltorange!50}{.508} & 23.39 & \cellcolor{pltorange!50}{.809} & \cellcolor{pltyellow!50}{.3466} \\
        
        NDR-Restore~\cite{yao2024neural} & \ding{55} & 196G & 36.9M & 23.84 & \cellcolor{pltorange!50}{.894} & 24.25 & \cellcolor{pltorange!50}{.802} & 20.08 & \cellcolor{pltorange!50}{.892} & 25.62 & \cellcolor{pltorange!50}{.912} & 24.37 & \cellcolor{pltorange!50}{.897} & 22.94 & \cellcolor{pltorange!50}{.669} & 23.52 & \cellcolor{pltorange!50}{.846} & \cellcolor{pltyellow!50}{.3126} \\
        
        Gridformer~\cite{wang2024gridformer} & \ding{55} & 367G & 34M & 23.12 & \cellcolor{pltorange!50}{.898} & 25.82 & \cellcolor{pltorange!50}{.783} & 19.24 & \cellcolor{pltorange!50}{.869} & 36.04 & \cellcolor{pltorange!50}{.937} & 31.72 & \cellcolor{pltorange!50}{.898} & 26.24 & \cellcolor{pltorange!50}{.623} & 27.03 & \cellcolor{pltorange!50}{.836} & \cellcolor{pltyellow!50}{.3754} \\
        
        DiffUIR-L~\cite{zheng2024selective} & \ding{55} & {10G} & 36.2M & 21.56 & \cellcolor{pltorange!50}{.812} & 23.85 & \cellcolor{pltorange!50}{.743} & 18.28 & \cellcolor{pltorange!50}{.864} & {36.84} & \cellcolor{pltorange!50}{{.938}} & {32.42} & \cellcolor{pltorange!50}{.897} & 26.08 & \cellcolor{pltorange!50}{.648} & 26.51 & \cellcolor{pltorange!50}{.818} & \cellcolor{pltyellow!50}{{.2564}} \\
        
        Histoformer~\cite{sun2025restoring} & \ding{55} & 91G & 16.6M & 23.22 & \cellcolor{pltorange!50}{.908} & 25.62 & \cellcolor{pltorange!50}{.782} & 19.78 & \cellcolor{pltorange!50}{{.903}} & 26.88 & \cellcolor{pltorange!50}{.845} & 25.64 & \cellcolor{pltorange!50}{.874} & 23.13 & \cellcolor{pltorange!50}{.659} & 24.04 & \cellcolor{pltorange!50}{.829} & \cellcolor{pltyellow!50}{.3524} \\

        adaIR~\cite{cui2024adair} & \ding{55} & 147G & 28.7M & 23.57 & \cellcolor{pltorange!50}{.916} & {26.35} & \cellcolor{pltorange!50}{{.801}} & 18.44 & \cellcolor{pltorange!50}{.901} & 32.84 & \cellcolor{pltorange!50}{.921} & 30.48 & \cellcolor{pltorange!50}{{.901}} & {26.48} & \cellcolor{pltorange!50}{{.672}} & 26.36 & \cellcolor{pltorange!50}{{.857}} & \cellcolor{pltyellow!50}{.3429} \\
        
        HAIR~\cite{cao2024hairhypernetworksbasedallinoneimage} & \ding{55} & 41G & 29M & {25.75} & \cellcolor{pltorange!50}{{.922}} & 25.78 & \cellcolor{pltorange!50}{.798} & {20.00} & \cellcolor{pltorange!50}{.894} & 35.54 & \cellcolor{pltorange!50}{.916} & 30.84 & \cellcolor{pltorange!50}{.878} & 26.26 & \cellcolor{pltorange!50}{.657} & {27.36} & \cellcolor{pltorange!50}{.847} & \cellcolor{pltyellow!50}{.2822} \\
        
        UHDprocesser~\cite{liu2025uhd} & \ding{51} & \textcolor{tabblue}{\textbf{4G}} & \textcolor{tabblue}{\textbf{1.6M}} & \textcolor{tabblue}{\textbf{27.11}} & \cellcolor{pltorange!50}\textcolor{tabblue}{\textbf{.925}} & \textcolor{tabblue}{\textbf{26.48}} & \cellcolor{pltorange!50}\textcolor{tabblue}{\textbf{.803}} & \textcolor{tabblue}{\textbf{20.94}} & \cellcolor{pltorange!50}\textcolor{tabblue}{\textbf{.923}} & \textcolor{tabblue}{\textbf{38.94}} & \cellcolor{pltorange!50}\textcolor{tabblue}{\textbf{.975}} & \textcolor{tabblue}{\textbf{33.99}} & \cellcolor{pltorange!50}\textcolor{tabblue}{\textbf{.903}} & \textcolor{tabblue}{\textbf{27.95}} & \cellcolor{pltorange!50}\textcolor{tabblue}{\textbf{.677}} & \textcolor{tabblue}{\textbf{29.23}} & \cellcolor{pltorange!50}\textcolor{tabblue}{\textbf{.868}} & \cellcolor{pltyellow!50}\textcolor{tabblue}{\textbf{.2541}} \\

        Ours & \ding{51} & \textcolor{tabred}{\textbf{3.6G}} & \textcolor{tabred}{\textbf{1.2M}} & \textcolor{tabred}{\textbf{27.32}} & \cellcolor{pltorange!50}\textcolor{tabred}{\textbf{.926}} & \textcolor{tabred}{\textbf{26.98}} & \cellcolor{pltorange!50}\textcolor{tabred}{\textbf{.811}} & \textcolor{tabred}{\textbf{21.21}} & \cellcolor{pltorange!50}\textcolor{tabred}{\textbf{.924}} & \textcolor{tabred}{\textbf{39.21}} & \cellcolor{pltorange!50}\textcolor{tabred}{\textbf{.978}} & \textcolor{tabred}{\textbf{34.78}} & \cellcolor{pltorange!50}\textcolor{tabred}{\textbf{.918}} & \textcolor{tabred}{\textbf{28.72}} & \cellcolor{pltorange!50}\textcolor{tabred}{\textbf{.707}} & \textcolor{tabred}{\textbf{29.70}} & \cellcolor{pltorange!50}\textcolor{tabred}{\textbf{.877}} & \cellcolor{pltyellow!50}\textcolor{tabred}{\textbf{.2502}} \\
        
        \Xhline{1pt}
    \end{tabularx}
\end{table*}

\begin{table*}[!t]
    \centering
    \tiny
    \fboxsep0.75pt
    \setlength\tabcolsep{3pt}
    \caption{\textit{Comparison to state-of-the-art on six degradations.} PSNR (dB, $\uparrow$), \colorbox{pltorange!50}{SSIM ($\uparrow$)}, \colorbox{pltyellow!50}{LPIPS ($\downarrow$)} and FS represents full-size 4K image inference. FLOPs are computed for an input size of 256$\times$256. \textcolor{tabred}{\textbf{Best}} and \textcolor{tabblue}{\textbf{second best}} performances are highlighted.}
    \label{tab:exp:6deg}
    \begin{tabularx}{0.97\textwidth}{Xc*{17}{c}}
    \Xhline{1pt}
     \multirow{2}{*}{Method} & \multirow{2}{*}{FS} & \multirow{2}{*}{\centering FLOPs} & \multirow{2}{*}{\centering Params.}  
     & \multicolumn{2}{c}{\textit{Low Light}} & \multicolumn{2}{c}{\textit{Deblurring}} & \multicolumn{2}{c}{\textit{Dehazing}} & \multicolumn{2}{c}{\textit{Denoising}} 
     &\multicolumn{2}{c}{\textit{Deraining}} & \multicolumn{2}{c}{\textit{Desnowing}} & \multicolumn{3}{c}{\multirow{2}{*}{\centering Average}} \\
     \cmidrule(lr){5-6} \cmidrule(lr){7-8} \cmidrule(lr){9-10} \cmidrule(lr){11-12} \cmidrule(lr){13-14} \cmidrule(lr){15-16}
     &&&& \multicolumn{2}{c}{UHD-LL} & \multicolumn{2}{c}{UHD-blur} & \multicolumn{2}{c}{UHD-haze} & \multicolumn{2}{c}{UHDN\textsubscript{$\sigma$=50}} & \multicolumn{2}{c}{UHD-rain} & \multicolumn{2}{c}{UHD-snow} \\
     \midrule
        AIRNet~\cite{AirNet} & \ding{55} & 301G & {9M} & 22.68 & \cellcolor{pltorange!50}{.887} & 23.52 &  \cellcolor{pltorange!50}{.876} & 18.24 & \cellcolor{pltorange!50}{.846} & 22.38 &  \cellcolor{pltorange!50}{.876} & 26.35 &  \cellcolor{pltorange!50}{.876} & 27.38 &  \cellcolor{pltorange!50}{.924} & 23.43 &  \cellcolor{pltorange!50}{.874} & \cellcolor{pltyellow!50}{.1861} \\

        IDR~\cite{zhang2023ingredient} & \ding{55} & 88G & 15.3M & 24.33 & \cellcolor{pltorange!50}{.915} & 25.64 &  \cellcolor{pltorange!50}{.788} & 18.68 &  \cellcolor{pltorange!50}{.879} & 29.64 &  \cellcolor{pltorange!50}{.906} & 28.82 &  \cellcolor{pltorange!50}{.906} & 30.48 &  \cellcolor{pltorange!50}{.945} & 26.27 &  \cellcolor{pltorange!50}{.890} & \cellcolor{pltyellow!50}{.1912} \\

        PromptIR~\cite{potlapalli2023promptir} & \ding{55} & 158G & 33M& 23.3 & \cellcolor{pltorange!50}{.911} & 26.48 & \cellcolor{pltorange!50}{{.805}} & 20.14 & \cellcolor{pltorange!50}{.901} & 24.88 & \cellcolor{pltorange!50}{.835} & 28.89 & \cellcolor{pltorange!50}{.897} & 30.78 & \cellcolor{pltorange!50}{.966} & 25.74 &  \cellcolor{pltorange!50}{.886} & \cellcolor{pltyellow!50}{.2155} \\

        CAPTNet~\cite{gao2023prompt} & \ding{55} & 25G & 24.3M & 24.97 & \cellcolor{pltorange!50}{{.921}} & {26.32} &  \cellcolor{pltorange!50}{.796} & {20.32} &  \cellcolor{pltorange!50}{.903} & 21.64 &  \cellcolor{pltorange!50}{.569} & {29.34} &  \cellcolor{pltorange!50}{{.908}} & {32.21} &  \cellcolor{pltorange!50}{{.974}} & 25.80 &  \cellcolor{pltorange!50}{.845} & \cellcolor{pltyellow!50}{.2861} \\
        
        NDR-Restore~\cite{yao2024neural} & \ding{55} & 196G & 36.9M & 25.12 & \cellcolor{pltorange!50}{.885} & 25.64 &  \cellcolor{pltorange!50}{.791} & 19.21 &  \cellcolor{pltorange!50}{.896} & 31.44 &  {{\cellcolor{pltorange!50}{.915}}} & 29.24 &   {{\cellcolor{pltorange!50}{.897}}} & 28.41 &   {{\cellcolor{pltorange!50}{.948}}} & 26.51 &  \cellcolor{pltorange!50}{.889} & \cellcolor{pltyellow!50}{.3108} \\

        Gridformer~\cite{wang2024gridformer} & \ding{55} & 367G & 34M & 23.92 & \cellcolor{pltorange!50}{.898} & 25.68 &  \cellcolor{pltorange!50}{.782} & 18.87 &  \cellcolor{pltorange!50}{.889} & 32.86 &  \cellcolor{pltorange!50}{.915} & 29.37 &  \cellcolor{pltorange!50}{.904} & 28.24 &  \cellcolor{pltorange!50}{.942} & 26.49&  \cellcolor{pltorange!50}{.895} & \cellcolor{pltyellow!50}{.2321} \\
        
        DiffUIR-L~\cite{zheng2024selective} & \ding{55} & {10G} & 36.2M & 22.64 & \cellcolor{pltorange!50}{.902} &  {{25.08}} &   \cellcolor{pltorange!50}{.785} &  {{18.62}} &   {{\cellcolor{pltorange!50}{.889}}} & {{33.25}} &  {\cellcolor{pltorange!50}{{.928}}} & {{27.89}} &  {\cellcolor{pltorange!50}{.886}} & {{27.36}} &  {\cellcolor{pltorange!50}{.945}} &  {{25.81}} &   {\cellcolor{pltorange!50}{.889}} & \cellcolor{pltyellow!50}{{.1844}} \\
        
        Histoformer~\cite{sun2025restoring} & \ding{55} & 91G & 16.6M & 25.73 & \cellcolor{pltorange!50}{.915} &  {{26.55}} &   \cellcolor{pltorange!50}{.796} &  {{18.73}} &   {{\cellcolor{pltorange!50}{.897}}} & {33.05} &  {\cellcolor{pltorange!50}{.924}} & {{27.96}} &  {\cellcolor{pltorange!50}{.884}} & {{27.56}} &  {\cellcolor{pltorange!50}{.971}} &  {{{26.59}} }&   {\cellcolor{pltorange!50}{.898}} & \cellcolor{pltyellow!50}{.1855} \\
        
        adaIR~\cite{cui2024adair} & \ding{55} & 147G & 28.7M & 23.84 & \cellcolor{pltorange!50}{.918} &  {{26.86}} &   \cellcolor{pltorange!50}{.803} &  {{19.34}} &   {{\cellcolor{pltorange!50}{{.910}}}} & {32.46} &  {\cellcolor{pltorange!50}{.923}} & {{28.18}} &  {\cellcolor{pltorange!50}{{.901}}} & {{27.72}} &  {\cellcolor{pltorange!50}{.953}} &  {{26.40}} &   {\cellcolor{pltorange!50}{.901}} & \cellcolor{pltyellow!50}{.2492} \\

        HAIR~\cite{cao2024hairhypernetworksbasedallinoneimage} & \ding{55} & 41G & 29M & {25.22} & \cellcolor{pltorange!50}{.897} &  {{24.77}} &   \cellcolor{pltorange!50}{.799} &  {{18.75}} &   {{\cellcolor{pltorange!50}{.883}}} & {32.50} &  {\cellcolor{pltorange!50}{.915}} & {{28.76}} &  {\cellcolor{pltorange!50}{.893}} & {{27.89}} &  {\cellcolor{pltorange!50}{.968}} &  {{26.31}} &   {\cellcolor{pltorange!50}{.892}} & \cellcolor{pltyellow!50}{.2607} \\

        UHDprocesser~\cite{liu2025uhd} & \ding{51} & \textcolor{tabblue}{\textbf{4G}} & \textcolor{tabblue}{\textbf{1.6M}} & \textcolor{tabblue}{\textbf{26.91}} & \textcolor{tabblue}{\textbf{\cellcolor{pltorange!50}\textcolor{tabblue}{{\textbf{.924}}}}} & \textcolor{tabblue}{\textbf{26.95}} & \textcolor{tabblue}{\textbf{\cellcolor{pltorange!50}{.807}}} & \textcolor{tabblue}{\textbf{21.81}} & \textcolor{tabblue}{\textbf{\cellcolor{pltorange!50}{.931}}} & \textcolor{tabblue}{\textbf{33.73}} & \textcolor{tabblue}{\textbf{\cellcolor{pltorange!50}{.934}}} & \textcolor{tabblue}{\textbf{29.90}} & \textcolor{tabblue}{\textbf{\cellcolor{pltorange!50}{.915}}} & \textcolor{tabblue}{\textbf{32.73}} & \textcolor{tabblue}{\textbf{\cellcolor{pltorange!50}{.979}}} & \textcolor{tabblue}{\textbf{28.67}} & \textcolor{tabblue}{\textbf{\cellcolor{pltorange!50}{.915}}} & \textcolor{tabblue}{\textbf{\cellcolor{pltyellow!50}{.1839}}} \\

        Ours & \ding{51} & \textcolor{tabred}{\textbf{3.6G}} & \textcolor{tabred}{\textbf{1.2M}} & \textcolor{tabred}{\textbf{27.14}} & \cellcolor{pltorange!50}\textcolor{tabred}{\textbf{.925}} & \textcolor{tabred}{\textbf{27.21}} & \cellcolor{pltorange!50}\textcolor{tabred}{\textbf{.815}} & \textcolor{tabred}{\textbf{22.32}} & \cellcolor{pltorange!50}\textcolor{tabred}{\textbf{.936}} & \textcolor{tabred}{\textbf{34.17}} & \cellcolor{pltorange!50}\textcolor{tabred}{\textbf{.942}} & \textcolor{tabred}{\textbf{31.41}} & \cellcolor{pltorange!50}\textcolor{tabred}{\textbf{.919}} & \textcolor{tabred}{\textbf{33.24}} & \cellcolor{pltorange!50}\textcolor{tabred}{\textbf{.982}} & \textcolor{tabred}{\textbf{29.24}} & \cellcolor{pltorange!50}\textcolor{tabred}{\textbf{.920}} & \cellcolor{pltyellow!50}\textcolor{tabred}{\textbf{.1822}} \\
        
        \Xhline{1pt}
    \end{tabularx}
    
\end{table*}
\begin{table*}[h]
\centering
\tiny
\caption{Adaptability in Standard-Resolution Scenarios.Comparisons use LPIPS and FID scores, with lower values indicating superior performance.}
\label{tab:Standard-Resolution}
\begin{tabular}{llcccccc}
\toprule
\textbf{Type} & \textbf{Method} & \textbf{Haze} & \textbf{Rain} & \textbf{Snow} & \textbf{Motion Blur} & \textbf{Raindrop} & \textbf{Low-light} \\
\midrule
Discriminative-based 
& PromptIR~\cite{potlapalli2023promptir}         & 0.309/141.05 & 0.097/32.61 & 0.100/18.34 & 0.163/35.79 & 0.189/84.48 & 0.421/189.87 \\
         & PromptIR /w Ours   & 0.224/121.12 & 0.086/28.68 & 0.092/17.12 & 0.161/35.12 & 0.182/76.84 & 0.378/172.59 \\
        
\midrule
LDM-based 

         & Diff-Plugin~\cite{liu2024diff}      & 0.340/143.66 & 0.165/39.71 & 0.178/18.08 & 0.147/37.68 & 0.185/60.64 & 0.466/167.63 \\
         & Diff-Plugin/w Ours                  & 0.321/131.12 & 0.162/39.43 & 0.174/18.02 & 0.138/35.42 & 0.146/44.26 & 0.432/152.28 \\
\midrule
VAE-based  
         
             & CosAE~\cite{liu2024cosae}     & 0.328/148.78 & 0.146/38.27 & 0.162/16.78 & 0.186/41.28 & 0.182/49.27 & 0.482/182.24 \\
             & CosAE/w Ours                  & 0.224/128.12 & 0.098/28.79 & 0.121/11.56 & 0.168/36.22 & 0.119/40.62 & 0.382/159.83 \\
\bottomrule
\end{tabular}
\end{table*}

\begin{table*}[!t] 
    \centering
    \tiny 
    \fboxsep0.75pt 
    \setlength\tabcolsep{3pt} 

    \vspace{-1em}
    \caption{Generalization Verification. PSNR (dB, $\uparrow$), \colorbox{pltorange!50}{SSIM ($\uparrow$)}, and \colorbox{pltyellow!50}{LPIPS ($\downarrow$)} are reported.}
    \label{tab:generalization_verification} 

    \begin{tabularx}{1\textwidth}{X *{18}{c}}
    \Xhline{1pt} 

    \multirow{2}{*}{Method} & \multicolumn{9}{c}{\textit{Unseen}} & \multicolumn{9}{c}{\textit{Composite Degradation}} \\
    \cmidrule(lr){2-10} \cmidrule(lr){11-19}

    & \multicolumn{3}{c}{UHD-rain} & \multicolumn{3}{c}{UHD-snow} & \multicolumn{3}{c}{UHD-moire} & \multicolumn{3}{c}{LLIE+Noise} & \multicolumn{3}{c}{Haze+LLIE} & \multicolumn{3}{c}{Noise+Blur} \\
    \midrule 

    HAIR & 24.32 & \cellcolor{pltorange!50}{0.924} & \cellcolor{pltyellow!50}{0.392} & 
             25.43 & \cellcolor{pltorange!50}{0.903} & \cellcolor{pltyellow!50}{0.223} & 
             17.28 & \cellcolor{pltorange!50}{0.798} & \cellcolor{pltyellow!50}{0.446} & 
             18.12 & \cellcolor{pltorange!50}{0.812} & \cellcolor{pltyellow!50}{0.492} & 
             16.72 & \cellcolor{pltorange!50}{0.862} & \cellcolor{pltyellow!50}{0.439} & 
             18.42 & \cellcolor{pltorange!50}{0.854} & \cellcolor{pltyellow!50}{0.471} \\ 

    UHD-processer & 22.72 & \cellcolor{pltorange!50}{0.924} & \cellcolor{pltyellow!50}{0.342} &
                    21.82 & \cellcolor{pltorange!50}{0.918} & \cellcolor{pltyellow!50}{0.267} &
                    14.32 & \cellcolor{pltorange!50}{0.778} & \cellcolor{pltyellow!50}{0.489} &
                    13.28 & \cellcolor{pltorange!50}{0.842} & \cellcolor{pltyellow!50}{0.428} &
                    12.38 & \cellcolor{pltorange!50}{0.872} & \cellcolor{pltyellow!50}{0.462} &
                    18.28 & \cellcolor{pltorange!50}{0.824} & \cellcolor{pltyellow!50}{0.492} \\

    Ours & \textcolor{tabred}{\textbf{28.13}} & \cellcolor{pltorange!50}\textcolor{tabred}{\textbf{0.892}} & \cellcolor{pltyellow!50}\textcolor{tabred}{\textbf{0.233}} &
           \textcolor{tabred}{\textbf{28.92}} & \cellcolor{pltorange!50}\textcolor{tabred}{\textbf{0.967}} & \cellcolor{pltyellow!50}\textcolor{tabred}{\textbf{0.184}} &
           \textcolor{tabred}{\textbf{19.26}} & \cellcolor{pltorange!50}\textcolor{tabred}{\textbf{0.898}} & \cellcolor{pltyellow!50}\textcolor{tabred}{\textbf{0.326}} &
           \textcolor{tabred}{\textbf{20.33}} & \cellcolor{pltorange!50}\textcolor{tabred}{\textbf{0.882}} & \cellcolor{pltyellow!50}\textcolor{tabred}{\textbf{0.342}} &
           \textcolor{tabred}{\textbf{19.82}} & \cellcolor{pltorange!50}\textcolor{tabred}{\textbf{0.904}} & \cellcolor{pltyellow!50}\textcolor{tabred}{\textbf{0.328}} &
           \textcolor{tabred}{\textbf{24.28}} & \cellcolor{pltorange!50}\textcolor{tabred}{\textbf{0.898}} & \cellcolor{pltyellow!50}\textcolor{tabred}{\textbf{0.278}} \\
    
    \Xhline{1pt} 
    \end{tabularx}
    \vspace{-1em}
\end{table*}

\subsection{Inference-Time Control}

The differentiated LoRA modules, trained with distinct losses and alternating optimization—$\Delta\phi_{LoRA}$ for fidelity extraction and $\Delta\psi_{LoRA}$ for perceptual generation—provide flexibility during inference. Users can introduce a control parameter $\alpha \in [0,1]$ to dynamically adjust the contributions of the encoder and decoder LoRA modules to the final result, for instance, via $\phi = \phi^* + \alpha \Delta\phi_{LoRA}$ and $\psi = \psi^* + (1-\alpha) \Delta\psi_{LoRA}$. This mechanism enables a tailored trade-off between maximizing fidelity and optimizing perceptual quality, depending on application requirements.

\vspace{-0.5em}
\section{Experiments}
\vspace{-0.5em}

\vspace{-0.5em}
\subsection{Ablation Studies}
\vspace{-0.5em}
\begin{table}[h]
\captionsetup[subtable]{font=scriptsize}
\vspace{-1em}
\caption{Ablation Studies.Comprehensive ablation experiments validate the efficacy of our approach.}
\begin{minipage}[]{0.48\textwidth}
    \centering
    \begin{subtable}{\linewidth}
        \tiny
        \begin{minipage}{\linewidth}
            \centering
            \label{tab:ablation}
            \caption{Ablation study of Latent Harmony.}
            
            \vspace{-1em}
        \end{minipage}
        \begin{tabular}{lccc}
            \toprule
            \textbf{Configuration} & \textbf{PSNR $\uparrow$} & \textbf{SSIM $\uparrow$} & \textbf{LPIPS $\downarrow$} \\
            \midrule
            Latent Harmony & \textbf{29.77} & \textbf{0.88} & \textbf{0.250} \\
            \midrule
            w/o $L_{\text{Inv}}$ & 24.28 & 0.79 & 0.292 \\
            w/o $L_{\text{Eqv}}$ & 25.68 & 0.82 & 0.302 \\
            w/o PDPS & 27.82 & 0.84 & 0.287 \\
            w/o FHF-LoRA & 28.12 & 0.86 & 0.286 \\
            w/o PHF-LoRA & 29.02 & 0.84 & 0.306 \\
            w/o LoRA Fine-Tuning & 28.68 & 0.85 & 0.298 \\
            w/o Fine-Tuning & 28.48 & 0.86 & 0.292 \\
            \bottomrule
        \end{tabular}
    \end{subtable}
    \begin{subtable}{\linewidth}
        \tiny
        \begin{minipage}{\linewidth}
            \centering
            \caption{Inference time comparison.}
            \label{tab:uhd_inference_time}
            \vspace{-1em}
        \end{minipage}
        \begin{tabular}{ c c c c}
            \toprule
             \textbf{DreamUIR} & \textbf{Histformer} & \textbf{UHDprocesser} & \textbf{LH (Ours)} \\
            \midrule
             12.3 & 8.4 & 1.2 & \textbf{0.43} \\
            \bottomrule
        \end{tabular}
    \end{subtable}
\end{minipage}
\hfill
\begin{minipage}[]{0.48\textwidth}
    \centering
    \begin{subtable}{\linewidth}
        \tiny
        \vspace{-1em}
        \setlength{\tabcolsep}{3pt}
        \begin{minipage}{\linewidth}
            
            \centering
            \caption{Ablation Study of Latent Restoration Network.}
            \label{tab:performance}
            \vspace{-1em}
        \end{minipage}
        \begin{tabular}{@{}l *{3}{cc} @{}}
            \toprule
            \multirow{2}{*}{metric} &
            \multicolumn{2}{c}{Restormer} &
            \multicolumn{2}{c}{NAFNet} &
            \multicolumn{2}{c}{SFHformer} \\
            \cmidrule(lr){2-3} \cmidrule(lr){4-5} \cmidrule(lr){6-7}
            & Base & +Ours & Base & +Ours & Base & +Ours \\
            \midrule
            PSNR (dB) & 24.22 & 29.73/ \textcolor{red}{{\textbf{+5.51}}} & 24.63 & 29.68/ \textcolor{red}{{\textbf{+5.05}}} & 24.54 & 29.70/ \textcolor{red}{{\textbf{+5.16}}} \\
            Param (M) & 26.1 & 3.8/ \textcolor{red}{\textbf{-85}\%} & 29.1 & 1.9/ \textcolor{red}{\textbf{-93}\%} & 7.6 & 1.2/ \textcolor{red}{\textbf{-84}\%} \\
            FLOPS (G) & 140.9 & 6.2/ \textcolor{red}{\textbf{-95}\%} & 16.1 & 4.7/ \textcolor{red}{\textbf{-71}\%} & 51.0 & 3.6/ \textcolor{red}{\textbf{-93}\%} \\
            Runtime (s) & 8.8 & 0.62/ \textcolor{red}{\textbf{-92}\%} & 4.6 & 0.41/ \textcolor{red}{\textbf{-92}\%} & 5.2 & 0.43/ \textcolor{red}{\textbf{-92}\%} \\
            FS & \ding{55} & \ding{51} & \ding{55} & \ding{51} & \ding{55} & \ding{51} \\
            \bottomrule
        \end{tabular}
    \end{subtable}
    \vspace{-1em}
    \begin{subtable}{\linewidth}
        \tiny
        \vspace{-1em}
        \begin{minipage}{\linewidth}
            \centering
            \caption{Performance metrics under different $\alpha$ values.}
            \label{tab:performance_alpha}
            \vspace{-1em}
        \end{minipage}
        \begin{tabular}{l c c c c}
            \toprule
            \textbf{Metric} & \textbf{\boldmath$\alpha=0.2$} & \textbf{\boldmath$\alpha=0.4$} & \textbf{\boldmath$\alpha=0.6$} & \textbf{\boldmath$\alpha=0.8$} \\
            \midrule
            PSNR & 28.94 & 29.28 & \textbf{29.70} & 29.74 \\
            SSIM & 0.862 & 0.867 & \textbf{0.877} & 0.878 \\
            LPIPS & 0.2218 & 0.2483 & \textbf{0.2502} & 0.2904 \\
            User & 9.2 & 7.8 & \textbf{6.2} & 4.8 \\
            \bottomrule
        \end{tabular}
    \end{subtable}
\end{minipage}
\end{table}

To validate the contributions of the key components in our ``Latent Harmony'' framework, we conducted thorough ablation experiments on the UHD all-in-one restoration task, systematically removing or modifying individual components and assessing their impact on performance using PSNR, SSIM, and LPIPS metrics. Results are summarized in ~\cref{tab:ablation}. The ablation of primary components, presented in~\cref{tab:ablation}(a), confirms the effectiveness of each proposed element. Additionally, the latent space restoration network in Latent Harmony adopts SFHformer~\cite{WhenFast}, and we verify the robustness of our approach across alternative network architectures in~\cref{tab:ablation}(c). Runtime comparisons, shown in~\cref{tab:ablation}(b), demonstrate significant efficiency gains over competing methods, underscoring the necessity of eliminating degradation-aware branches. The impact of the tuning parameter $\alpha$ on fidelity and perceptual quality is illustrated in~\cref{tab:ablation}(d), where increasing $\alpha$ enhances fidelity metrics at the expense of perceptual metrics, validating the tunability of our method’s output. Detailed experimental setups, implementation specifics, and additional results and analyses are provided in supplementary.

\subsection{All-in-One and Single-Task Restoration on UHD Scenes}

\vspace{-0.5em}
We evaluated the efficacy of our proposed method for the UHD all-in-one restoration task across two experimental configurations: four-degradation and six-degradation settings. As reported in ~\cref{tab:exp:4deg,tab:exp:6deg}, our approach consistently achieved state-of-the-art performance in both settings while maintaining optimal computational efficiency. Fig.~\ref{fig:visual} shows visual results of the four-degradation setting, depicting that our method effectively removes degradations while preserving intricate background textures. Although our method is designed for all-in-one tasks, it does not significantly compromise single-task restoration performance, as detailed in the supplementary.
\vspace{-0.5em}
\subsection{Adaptability and Application Exploration on Standard-Resolution Scenes}
\vspace{-0.5em}

Our method employs a unified processing strategy for all degradations, eschewing specialized degradation-aware branches, thereby achieving superior generalization compared to traditional approaches. We validated this generalization capability under two experimental settings: unseen degradations excluded from training and novel composite degradations formed by combining trained degradation types. As shown in ~\cref{tab:generalization_verification}, our approach significantly enhances generalization performance in both scenarios, demonstrating that the homogeneous latent space processing paradigm proposed in this work is a more robust alternative to incorporating degradation-aware branches.

The primary objective of this work is to develop a VAE framework tailored for UHD restoration tasks. However, VAEs are also widely employed in standard-resolution scenarios to enhance perceptual quality and reduce the computational demands of diffusion-based restoration methods. To demonstrate the versatility of our approach, we integrated our proposed LH-VAE with three representative standard-resolution restoration methods: discriminative-based, Latent Diffusion Model (LDM)-based, and VAE-based. Experiments were conducted on a multi-degradation benchmark curated from the Gendeg dataset. As shown in~\cref{tab:Standard-Resolution}, our method significantly improves the perceptual metrics of all three approaches in standard-resolution settings, thereby validating its generalizability.

\vspace{-1em}
\section{Conclusion}
\vspace{-1em}
This paper addressed critical VAE-based trade-offs in UHD all-in-one image restoration, encompassing latent generalization versus reconstruction, structural integrity during co-optimization, and the perception-fidelity balance. We introduced \textit{Latent Harmony},"a two-stage framework. Its first stage constructs a robust Latent Harmony VAE (LH-VAE) via principled latent space regularization. The second stage features high-frequency-guided LoRA fine-tuning, distinctly optimizing encoder LoRA for fidelity and decoder LoRA for perception, while preserving VAE structure. An inference parameter $\alpha$ enables explicit fidelity-perception control. Extensive experiments validated \textit{Latent Harmony's} superior restoration performance and effective balancing of these trade-offs across diverse scenarios, presenting a promising advancement for UHD image restoration.

{
\small
\bibliographystyle{IEEEtran} \bibliography{neurips_2025}
}



\newpage
\section*{NeurIPS Paper Checklist}

The checklist is designed to encourage best practices for responsible machine learning research, addressing issues of reproducibility, transparency, research ethics, and societal impact. Do not remove the checklist: {\bf The papers not including the checklist will be desk rejected.} The checklist should follow the references and follow the (optional) supplemental material.  The checklist does NOT count towards the page
limit. 

Please read the checklist guidelines carefully for information on how to answer these questions. For each question in the checklist:
\begin{itemize}
    \item You should answer \answerYes{}, \answerNo{}, or \answerNA{}.
    \item \answerNA{} means either that the question is Not Applicable for that particular paper or the relevant information is Not Available.
    \item Please provide a short (1–2 sentence) justification right after your answer (even for NA). 
\end{itemize}

{\bf The checklist answers are an integral part of your paper submission.} They are visible to the reviewers, area chairs, senior area chairs, and ethics reviewers. You will be asked to also include it (after eventual revisions) with the final version of your paper, and its final version will be published with the paper.

The reviewers of your paper will be asked to use the checklist as one of the factors in their evaluation. While "\answerYes{}" is generally preferable to "\answerNo{}", it is perfectly acceptable to answer "\answerNo{}" provided a proper justification is given (e.g., "error bars are not reported because it would be too computationally expensive" or "we were unable to find the license for the dataset we used"). In general, answering "\answerNo{}" or "\answerNA{}" is not grounds for rejection. While the questions are phrased in a binary way, we acknowledge that the true answer is often more nuanced, so please just use your best judgment and write a justification to elaborate. All supporting evidence can appear either in the main paper or the supplemental material, provided in appendix. If you answer \answerYes{} to a question, in the justification please point to the section(s) where related material for the question can be found.

IMPORTANT, please:
\begin{itemize}
    \item {\bf Delete this instruction block, but keep the section heading ``NeurIPS Paper Checklist"},
    \item  {\bf Keep the checklist subsection headings, questions/answers and guidelines below.}
    \item {\bf Do not modify the questions and only use the provided macros for your answers}.
\end{itemize}


\begin{enumerate}

\item {\bf Claims}
    \item[] Question: Do the main claims made in the abstract and introduction accurately reflect the paper's contributions and scope?
    \item[] Answer: \answerYes{} 
    \item[] Justification:  We summarize our contributions at the end of the introduction.

\item {\bf Limitations}
    \item[] Question: Does the paper discuss the limitations of the work performed by the authors?
    \item[] Answer:\answerYes{} 
    \item[] Justification:  We explicitly list limitations of our work in appendix.

\item {\bf Theory assumptions and proofs}
    \item[] Question: For each theoretical result, does the paper provide the full set of assumptions and a complete (and correct) proof?
    \item[] Answer: \answerYes{} 
    \item[] Justification:  It is explained in the motivation and method.

    \item {\bf Experimental result reproducibility}
    \item[] Question: Does the paper fully disclose all the information needed to reproduce the main experimental results of the paper to the extent that it affects the main claims and/or conclusions of the paper (regardless of whether the code and data are provided or not)?
    \item[] Answer:  \answerYes{}
    \item[] Justification: We have included implementation details in appendix.

\item {\bf Open access to data and code}
    \item[] Question: Does the paper provide open access to the data and code, with sufficient instructions to faithfully reproduce the main experimental results, as described in supplemental material?
    \item[] Answer:\answerNo{} 
    \item[] Justification:We will release the source code upon acceptance of the paper.

\item {\bf Experimental setting/details}
    \item[] Question: Does the paper specify all the training and test details (e.g., data splits, hyperparameters, how they were chosen, type of optimizer, etc.) necessary to understand the results?
    \item[] Answer: \answerYes{} 
    \item[] Justification:  It is explained in appendix.

\item {\bf Experiment statistical significance}
    \item[] Question: Does the paper report error bars suitably and correctly defined or other appropriate information about the statistical significance of the experiments?
    \item[] Answer: \answerYes{} 
    \item[] Justification: We discussed the details in the paper and it is statistically meaningful.

\item {\bf Experiments compute resources}
    \item[] Question: For each experiment, does the paper provide sufficient information on the computer resources (type of compute workers, memory, time of execution) needed to reproduce the experiments?
    \item[] Answer: \answerYes{} 
    \item[] Justification: It is explained in the appendix.
    
\item {\bf Code of ethics}
    \item[] Question: Does the research conducted in the paper conform, in every respect, with the NeurIPS Code of Ethics \url{https://neurips.cc/public/EthicsGuidelines}?
    \item[] Answer:  \answerYes{} 
    \item[] Justification: Yes, we confirm.

\item {\bf Broader impacts}
    \item[] Question: Does the paper discuss both potential positive societal impacts and negative societal impacts of the work performed?
    \item[] Answer:\answerYes{} 
    \item[] Justification:  This is discussed in appendix.
    
\item {\bf Safeguards}
    \item[] Question: Does the paper describe safeguards that have been put in place for responsible release of data or models that have a high risk for misuse (e.g., pretrained language models, image generators, or scraped datasets)?
    \item[] Answer:\answerNo{} 
    \item[] Justification: \justificationTODO{}
    \item[] Guidelines: We believe there is no such risk.
    \begin{itemize}
        \item The answer NA means that the paper poses no such risks.
        \item Released models that have a high risk for misuse or dual-use should be released with necessary safeguards to allow for controlled use of the model, for example by requiring that users adhere to usage guidelines or restrictions to access the model or implementing safety filters. 
        \item Datasets that have been scraped from the Internet could pose safety risks. The authors should describe how they avoided releasing unsafe images.
        \item We recognize that providing effective safeguards is challenging, and many papers do not require this, but we encourage authors to take this into account and make a best faith effort.
    \end{itemize}

\item {\bf Licenses for existing assets}
    \item[] Question: Are the creators or original owners of assets (e.g., code, data, models), used in the paper, properly credited and are the license and terms of use explicitly mentioned and properly respected?
    \item[] Answer:  \answerYes{} 
    \item[] Justification:  We credit the original owners of all assets.

\item {\bf New assets}
    \item[] Question: Are new assets introduced in the paper well documented and is the documentation provided alongside the assets?
    \item[] Answer: \answerYes{} 
    \item[] Justification: We will release the source code and model upon acceptance of the paper.

\item {\bf Crowdsourcing and research with human subjects}
    \item[] Question: For crowdsourcing experiments and research with human subjects, does the paper include the full text of instructions given to participants and screenshots, if applicable, as well as details about compensation (if any)? 
    \item[] Answer: \answerNA{} 
    \item[] Justification: \answerNA{}

\item {\bf Institutional review board (IRB) approvals or equivalent for research with human subjects}
    \item[] Question: Does the paper describe potential risks incurred by study participants, whether such risks were disclosed to the subjects, and whether Institutional Review Board (IRB) approvals (or an equivalent approval/review based on the requirements of your country or institution) were obtained?
    \item[] Answer: \answerNA{} 
    \item[] Justification: \answerNA{}

\item {\bf Declaration of LLM usage}
    \item[] Question: Does the paper describe the usage of LLMs if it is an important, original, or non-standard component of the core methods in this research? Note that if the LLM is used only for writing, editing, or formatting purposes and does not impact the core methodology, scientific rigorousness, or originality of the research, declaration is not required.
    \item[] Answer: \answerNA{} 
    \item[] Justification: \answerNA{}

\end{enumerate}

\end{document}